\documentclass[11pt,a4paper]{article}
\usepackage[hyperref]{naaclhlt2018}
\usepackage{times}
\usepackage{latexsym}
\usepackage{graphicx}
\usepackage{url}
\usepackage{amsmath}

\aclfinalcopy 

\title{Luminoso at SemEval-2018 Task 10: Distinguishing Attributes Using Text Corpora and Relational Knowledge}

\author{Robyn Speer \\
  Luminoso Technologies, Inc. \\
  675 Massachusetts Avenue \\
  Cambridge, MA 02139 \\
  {\tt rspeer@luminoso.com} \\\And
  Joanna Lowry-Duda \\
  Luminoso Technologies, Inc. \\
  675 Massachusetts Avenue \\
  Cambridge, MA 02139 \\
  {\tt jlowry-duda@luminoso.com} \\}

\date{}

\begin{document}
\maketitle
\begin{abstract}

    Luminoso participated in the SemEval 2018 task on ``Capturing
    Discriminative Attributes'' with a system based on ConceptNet, an open
    knowledge graph focused on general knowledge. In this paper, we describe
    how we trained a linear classifier on a small number of
    semantically-informed features to achieve an $F_1$ score of 0.7368 on the task,
    close to the task's high score of 0.75.

\end{abstract}

\newcommand{\term}[0]{$\mathit{term}$}
\newcommand{\termOne}[0]{$\mathit{term}_1$}
\newcommand{\termTwo}[0]{$\mathit{term}_2$}
\newcommand{\att}[0]{$\mathit{att}$}

\section{Introduction}

Word embeddings are most effective when they learn from both unstructured text
and a graph of general knowledge \cite{speer-lowryduda:2017:SemEval}. ConceptNet 5
\cite{speer2017conceptnet} is an open-data knowledge graph that is well suited for
this purpose. It is accompanied by a pre-built word embedding model known as
ConceptNet
Numberbatch\footnote{\url{https://github.com/commonsense/conceptnet-numberbatch}},
which combines skip-gram embeddings learned from unstructured text with the
relational knowledge in ConceptNet.

A straightforward application of the ConceptNet Numberbatch embeddings took first
place in SemEval 2017 task 2, on semantic word similarity. For SemEval 2018, we
built a system with these embeddings as a major component for a slightly more
complex task.

The Capturing Discriminative Attributes task \cite{semeval2018task10} emphasizes
the ability of a semantic model to recognize relevant differences between
terms, not just their similarities. As the task description states, ``If you
can tell that americano is similar to capuccino and espresso but you can't tell
the difference between them, you don't know what americano is.''

The ConceptNet Numberbatch embeddings only measure the similarity of terms,
and we hypothesized that we would need to represent more specific relationships.
For example, the input triple ``frog, snail, legs'' asks us to determine whether
``legs'' is an attribute that distinguishes ``frog'' from ``snail''. The answer
is yes, because a frog \emph{has} legs while a snail does not.
The \emph{has} relationship is one example of a specific relationship that is
represented in ConceptNet.

To capture this kind of specific relationship, we built a model that infers
relations between ConceptNet nodes, trained on the existing edges in ConceptNet
and random negative examples. There are many models designed for this purpose;
the one we decided on is based on Semantic Matching Energy (SME)
\cite{bordes2014semantic}.

Our features consisted of direct similarity over ConceptNet Numberbatch embeddings, the
relationships inferred over ConceptNet by SME, features that compose ConceptNet
with other resources (WordNet and Wikipedia), and a purely corpus-based feature
that looks up two-word phrases in the Google Books dataset.

We combined these features based on ConceptNet with features extracted from a
few other resources in a LinearSVC classifier, using liblinear
\cite{fan2008liblinear} via scikit-learn \cite{pedregosa2011scikit}. The
classifier used only 15 features, of which 12 ended up with non-zero weights,
from the five sources described. We aimed to avoid complexity in the classifier
in order to prevent overfitting to the validation set; the power of the
classifier should be in its features.

The classifier produced by this design (submitted late to the contest
leaderboard) successfully avoided overfitting. It performed better on the test
set than on the validation set, with a test $F_1$ score of 0.7368, whose margin of
error overlaps with the evaluation's reported high score of 0.75.

At evaluation time, we accidentally submitted our results on the validation
data, instead of the test data, to the SemEval leaderboard. Our code had
truncated the results to the length of the test data, causing us to not notice
the mismatch. This erroneous submission got a very low score, of course. This
paper presents the corrected test results, which we submitted to the
post-evaluation CodaLab leaderboard immediately after the results appeared.  We
did not change the classifier or data; the change was a one-line change to our
code for outputting the classifier's predictions on the test set instead on the
validation set.

\section{Features}

In detail, these are the five sources of features we used:

\paragraph{ConceptNet vector similarity.} Given the triple (\termOne, \termTwo,
\att), we look up the ConceptNet Numberbatch embeddings for the root words of the
three terms (with root words determined using ConceptNet's built-in
lemmatizer). We determine the cosine similarity of (\termOne, \att) and the
cosine similarity of (\termTwo, \att). We then subtract the square roots of the
similarity scores (floored at 0). If this difference is large enough, it
indicates a positive example, a discriminative attribute that applies to
\termOne{} and not to \termTwo.

\paragraph{ConceptNet relational inference.} We train a Semantic
Matching Energy model to represent ConceptNet nodes and relations as vectors,
along with a 3-tensor of interactions between them. This model can then assign
a confidence score to any triple (a relation connecting two terms). We used
this model to infer values for each of 11 different ConceptNet relations.  As
in the case of vector similarity, each feature value is the difference between
the value inferred for \emph{rel}(\termOne, \att) and \emph{rel}(\termTwo,
\att). This model is described in more detail in the next section.

\paragraph{Wikipedia lead sections.} This feature expands on ConceptNet vector
similarity: instead of computing the similarity between the attribute and the
term, it computes the maximum of the similarity between the attribute and any
word that appears in the lead section of the Wikipedia article for the term
\cite{wikipedia2017en}. This helps to identify attributes that would be used to
define the term, such as ``amphibian'' as an attribute for ``frog''.

\paragraph{WordNet entries.} This feature is similar to the ``Wikipedia lead
sections'' feature.  It expands each term by looking up its synonyms in WordNet
\cite{miller1998wordnet}, the synonyms in synsets it is connected to, and the
words in its gloss (definition), and taking the maximum similarity of the
attribute to any of these terms.

\paragraph{Google Books 2-grams.} This feature determines if \termOne{} forms a
significant two-word phrase with \att, more than \termTwo{} does, based on the
Google Books English Fiction data \cite{lin2012ngrams}.
The ``significance'' (s) of a two-word phrase is determined by comparing the
smoothed log-likelihood of the individual unigrams to the smoothed
log-likelihood of the phrase:
\begin{equation*}
    \begin{split}
        \mathrm{s}(\mathit{term}, \mathit{att}) = 10 + \log_{10}(\#(\mathit{term}, \mathit{att}) + 1)\\
        - \log_{10}((\#(\mathit{term}) + 10^5)(\#(\mathit{att}) + 10^5))
    \end{split}
\end{equation*}
where $\#$ represents the number of occurrences of a unigram or bigram in the corpus.

The ``ConceptNet relational inference'' feature provides 11 entries to the
feature vectors, while the other sources each provide one. In total, there are 15
features that represent each input triple.

Across multiple data sources, we use the square root of cosine similarity to
measure the strength of the match between a term and an attribute.
Because attributes should be at least somewhat related to the terms they
describe, and because weak semantic similarity can be interpreted as relatedness,
the square root helps us emphasize the important part of the scale. The difference
between ``somewhat related'' and ``not related'' is more important to the task
than the difference between ``very similar'' and ``somewhat related'', as a
discriminative attribute should ideally be unrelated to the second term.

\subsection{The Relational Inference Model}

To infer truth values for ConceptNet relations, we use a variant of the
Semantic Matching Energy model \cite{bordes2014semantic}, adapted to work well on
ConceptNet's vocabulary of relations. Instead of embedding relations in the
same space as the terms, this model assigns new 10-dimensional embeddings
to ConceptNet relations, yielding a compact model for ConceptNet's relatively
small set of relations.

The model is trained to distinguish positive examples of ConceptNet edges
from negative ones. The positive examples are edges directly contained in
ConceptNet, or those that are entailed by changing the relation to a more
general one or switching the directionality of a symmetric relation. The
negative examples come from replacing one of the terms with a random other
term, the relation with a random unentailed relation, or switching the
directionality of an asymmetric relation.

We trained this model for approximately 3 million iterations (about 4 days of
computation on an nVidia Titan Xp) using PyTorch \cite{paszke2017automatic}.
The code of the model is available at
\url{https://github.com/LuminosoInsight/conceptnet-sme}.

To extract features for the discriminative attribute task, we focus on a subset
of ConceptNet relations that would plausibly be used as attributes:
\emph{RelatedTo}, \emph{IsA}, \emph{HasA}, \emph{PartOf},
\emph{CapableOf}, \emph{UsedFor}, \emph{HasContext},
\emph{HasProperty}, and \emph{AtLocation}.

For most of these relations, the first argument is the term, and the second
argument is the attribute. We use two additional features for \emph{PartOf} and
\emph{AtLocation} with their arguments swapped, so that the attribute is the
first argument. The generic relation \emph{RelatedTo}, unlike the others, is
intended to be symmetric, so we add its value to the value of its swapped
version and use it as a single feature.

\section{The Overfitting-Resistant Classifier}

The classifier that we use to make a decision based on these features is
scikit-learn's LinearSVC, using the default parameters in scikit-learn 0.19.1.
(In Section~\ref{sec:experiments}, we discuss other models and parameters that
we tried.) This classifier makes effective use of the features while being
simple enough to avoid some amount of overfitting.

One aspect of the classifier that made a noticeable difference was the
scaling of the features. We tried $L_1$ and $L_2$-normalizing the columns of
the input matrix, representing the values of each feature, and decided on
$L_2$ normalization.

We took advantage of the design of our features and the asymmetry of the task
as a way to further mitigate overfitting. All of the features were designed to
identify a property that \termOne{} has and \termTwo{} does not, as is the case
for the discriminative examples, so they should all make a non-negative
contribution to a feature being discriminative. We can inspect the coefficients
of the features in the SVC's decision boundary. If any feature gets a negative
weight, it is likely a spurious result from overfitting to the training data.
So, after training the classifier, we clip the coefficients of the decision
boundary, setting all negative coefficients to zero.

If we were to remove these features and re-train, or require non-negative
coefficients as a constraint on the classifier, then other features would
inherently become responsible for overfitting. By neutralizing the features
\emph{after} training, we keep the features that are working well as they are,
and remove a part of the model that appears to purely represent overfitting.
Indeed, clipping the negative coefficients in this way increased our
performance on the validation set.

Table~\ref{table:coefficients} shows the coeffcients assigned to each feature
based on the training data.

\begin{table}[t]
\begin{small}
\begin{tabular}{lr}
\textbf{Feature} & \textbf{Coefficient}\\
\hline
ConceptNet vector similarity	& 13.82\\
SME: RelatedTo			& 14.01\\
SME: ($x$ IsA $a$)			& 2.13\\
SME: ($x$ HasA $a$)			& 0.00\\
SME: ($x$ PartOf $a$)		& 0.56\\
SME: ($x$ CapableOf $a$)		& 3.72\\
SME: ($x$ UsedFor $a$)		& 0.92\\
SME: ($x$ HasContext $a$)		& 0.88\\
SME: ($x$ HasProperty $a$)		& 0.00\\
SME: ($x$ AtLocation $a$)		& 0.00\\
SME: ($a$ PartOf $x$)		& 3.22\\
SME: ($a$ AtLocation $x$)		& 0.69\\
Wikipedia lead sections		& 12.46\\
WordNet relatedness		& 13.95\\
Google Ngrams			& 28.82\\
\end{tabular}
\end{small}
\caption{Coefficients of each feature in our linear classifier.
    $x$ represents a term and $a$ represents the attribute.}
\label{table:coefficients}
\end{table}

\section{Other experiments}

\label{sec:experiments}
There are other features that we tried and later discarded. We experimented
with a feature similar to the Google Books 2-grams feature, based on the AOL
query logs dataset \cite{pass2006}. It did not add to the performance, most
likely because any information it could provide was also provided by Google
Books 2-grams. Similiarly, we tried extending the Google Books 2-grams data to
include the first and third words of a selection of 3-grams, but this, too,
appeared redundant with the 2-grams.

We also experimented with a feature based on bounding box annotations available
in the OpenImages dataset \cite{openimages}. We hoped it would help us capture
attributes such as colors, materials, and shapes. While this feature did not
improve the classifier's performance on the validation set, it did slightly
improve the performance on the test set.

Before deciding on scikit-learn's LinearSVC, we experimented with a number of
other classifiers. This included random forests, differentiable models made of
multiple ReLU and sigmoid layers, and SVM with an RBF kernel or a polynomial
kernel.

We also experimented with different parameters to LinearSVC, such as changing the
default value of the penalty parameter $C$ of the error term, changing the
penalty from $L_2$ to $L_1$, solving the primal optimization problem instead of
the dual problem, and changing the loss from squared hinge to hinge. These
changes either led to lower performance or had no significant effect, so in the
end we used LinearSVC with the default parameters for scikit-learn version
0.19.1.

\section{Results}

\begin{table}[t]
\begin{small}
\begin{tabular}{lll}
\textbf{Dataset} & \textbf{F1} & \textbf{Error} (SEM)\\
\hline
train 		& .7617 & $\pm$ .0032\\
validation	& .7281 & $\pm$ .0085\\
test		& .7368 & $\pm$ .0091\\
\end{tabular}
\end{small}
\caption{$F_1$ scores by dataset. The reported $F_1$ score is the arithmetic
mean of the $F_1$ scores for both classes.}
\label{table:results-table}
\end{table}

When trained on the training set, the classifier we describe achieved an $F_1$
score of 0.7617 on the training set, 0.7281 on the validation set, and 0.7368
on the test set. Table~\ref{table:results-table} shows these scores along with
their standard error of the mean, supposing that these data sets were randomly
sampled from larger sets.

\subsection{Ablation Analysis}

\begin{figure}[t]
\centering
\includegraphics[width=75mm,trim=40 40 60 60,clip]{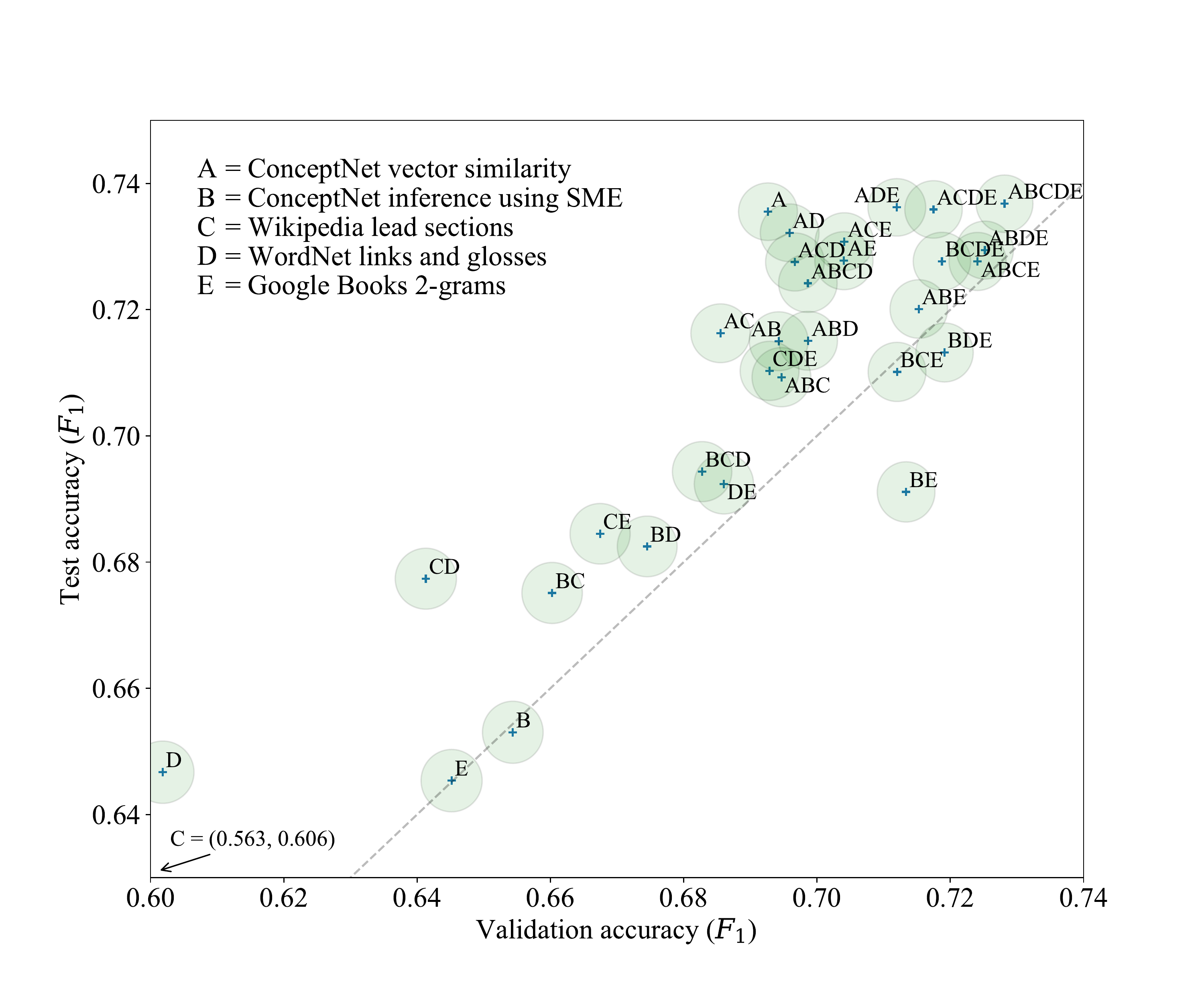}
\caption{This ablation analysis shows the contributions of subsets of the
    five sources of features. Ellipses indicate standard error of the mean,
    assuming that the data is sampled from a larger, unseen set.}
\label{fig:ablation}
\end{figure}

We performed an ablation analysis to see what the contribution of each of our
five sources of features was. We evaluated classifiers that used all non-empty
subsets of these sources. Figure~\ref{fig:ablation} plots the results of these
31 classifiers when evaluated on the validation set and the test set.

It is likely that the classifier with all five sources (\emph{ABCDE}) performed the best
overall. It is in a statistical tie ($p > .05$) with \emph{ABDE}, the classifier that
omits Wikipedia as a source.

Most of the classifiers perfomed better on the test set than on the validation
set, as shown by the dotted line. Some simple classifiers with very few
features performed particularly well on the test set. One surprisingly
high-performing classifier was \emph{A} (ConceptNet vector similarity), which
gets a test $F_1$ score of 0.7355 $\pm$ 0.0091. This is simple enough to be
called a heuristic instead of a classifier, and we can express it in closed
form. It is equivalent to this expression over ConceptNet Numberbatch
embeddings: $$ \mathrm{sim}(\mathit{term}_1, \mathit{att}) -
\mathrm{sim}(\mathit{term}_2, \mathit{att}) > 0.0961 $$ where $\mathrm{sim}(a,
b) = \sqrt{\max\left(\frac{a \cdot b}{||a|| \cdot ||b||}, 0\right)}$.

It is interesting to note that source \emph{A} (ConceptNet vector similarity)
appears to dominate source \emph{B} (ConceptNet SME) on the test data. SME led
to improvements on the validation set, but on the test set, any classifier
containing \emph{AB} performs equal to or worse than the same classifier with
\emph{B} removed. This may indicate that the SME features were the most prone
to overfitting, or that the validation set generally required making more
difficult distinctions than the test set.

\section{Reproducing These Results}

The code for our classifier is available on GitHub at
\url{https://github.com/LuminosoInsight/semeval-discriminatt}, and its
input data is downloadable from \url{https://zenodo.org/record/1183358}.

\bibliography{conceptnet}
\bibliographystyle{acl_natbib}

\end{document}